\documentclass[11pt,a4paper]{article}
\usepackage[hyperref]{acl2020}
\usepackage{times}
\usepackage{latexsym}
\usepackage{mathrsfs}
\usepackage{amsmath}
\usepackage{amssymb}
\usepackage{graphicx}
\usepackage{algorithm}
\usepackage{algpseudocode}
\usepackage{multirow}

\usepackage{microtype}

\aclfinalcopy 

\title{Meta-Transfer Learning for Code-Switched Speech Recognition}

\author{Genta Indra Winata$^{\dagger}$\thanks{\hspace{1mm} These two authors contributed equally.} , Samuel Cahyawijaya$^*$, Zhaojiang Lin, \\
\textbf{Zihan Liu}, \textbf{Peng Xu}, \textbf{Pascale Fung} \\
Center for Artificial Intelligence Research (CAiRE)\\
Department of Electronic and Computer Engineering \\
The Hong Kong University of Science and Technology\\
{$^\dagger$}\texttt{giwinata@connect.ust.hk}}

\date{}

\begin{document}
\maketitle
\begin{abstract}
An increasing number of people in the world today speak a mixed-language as a result of being multilingual. However, building a speech recognition system for code-switching remains difficult due to the availability of limited resources and the expense and significant effort required to collect mixed-language data. We therefore propose a new learning method, \textit{meta-transfer learning}, to transfer learn on a code-switched speech recognition system in a low-resource setting by judiciously extracting information from high-resource monolingual datasets. Our model learns to recognize individual languages, and transfer them so as to better recognize mixed-language speech by conditioning the optimization on the code-switching data. Based on experimental results, our model outperforms existing baselines on speech recognition and language modeling tasks, and is faster to converge.
\end{abstract}

\section{Introduction}
In bilingual or multilingual communities, speakers can easily switch between different languages within a conversation ~\cite{wang2009sustained}. 
People who know how to code-switch will mix languages in response to social factors as a way of communicating in a multicultural society. Generally, code-switching speakers switch languages by taking words or phrases from the embedded language to the matrix language. This can occur within a sentence, which is known as \textit{intra-sentential code-switching} or between two matrix language sentences, which is called \textit{inter-sentential code-switching}~\cite{heredia2001bilingual}.

Learning a code-switching automatic speech recognition (ASR) model has been a challenging task for decades due to data scarcity and difficulty in capturing similar phonemes in different languages. Several approaches have focused on generating synthetic speech data from monolingual resources~\cite{nakayama2018speech,winata2019code}. However, these methods are not guaranteed to generate natural code-switching speech or text. Another line of work explores the feasibility of leveraging large monolingual speech data in the pre-training and applying fine-tuning on the model using a limited source of code-switching data, which has been found useful to improve the performance~\cite{li2011asymmetric,winata2019code}. However, the transferability of these pretraining approaches is not optimized on extracting useful knowledge from each individual languages in the context of code-switching, and even after the fine-tuning step, the model forgets about the previously learned monolingual tasks.

\begin{figure}[t!]
    \centering
    \includegraphics[scale=1.35]{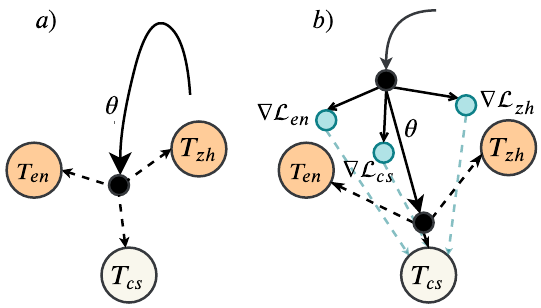}
    \caption{Illustration of (a) joint training and (b) meta-transfer learning. The solid lines show the optimization path. The orange circles represent the monolingual source language, and the white circles represent the code-switching target language. The lower black circle in (b) is closer to $T_{cs}$ than that in (a).}
    \label{fig:model}
\end{figure}

In this paper, we introduce a new method, \textit{meta-transfer learning}\footnote{The code is available at \href{https://github.com/audioku/meta-transfer-learning}{https://github.com/audioku/meta-transfer-learning}}, to learn to transfer knowledge from source monolingual resources to a code-switching model. Our approach extends the model-agnostic meta learning (MAML)~\cite{finn2017model} to not only train with monolingual source language resources but also optimize the update on the code-switching data. This allows the model to leverage monolingual resources that are optimized to detect code-switching speech. Figure~\ref{fig:model} illustrates the optimization flow of the model. Different from joint training, meta-transfer learning computes the first-order optimization using the gradients from monolingual resources constrained to the code-switching validation set. Thus, instead of learning one model that is able to generalize to all tasks, we focus on judiciously extracting useful information from the monolingual resources.

The main contribution is to propose a novel method to transfer learn information efficiently from monolingual resources to the code-switched speech recognition system. We show the effectiveness of our approach in terms of error rate, and that our approach is also faster to converge. We also show that our approach is also applicable to other natural language tasks, such as code-switching language modeling tasks.
 




\section{Related Work}
\paragraph{Meta-learning} Our idea of learning knowledge transfer from source monolingual resources to a code-switching model comes from MAML~\cite{finn2017model}. Probabilistic MAML~\cite{finn2018pmaml} is an extension of MAML, which has better classification coverage. Meta-learning has been applied to natural language and speech processing~\cite{hospedales2020meta}.~\citet{madotto2019paml} extends MAML to the personalized text generation domain and successfully produces more persona-consistent dialogue.
~\citet{gu2018meta} and \citet{qian-yu-2019-domain} propose to apply meta-learning on low-resource learning.~\citet{lin2019learning} applies MAML to low-resource sales prediction. Several applications have been proposed in speech applications, such as cross-lingual speech recognition~\cite{hsu2019meta}, speaker adaptation~\cite{klejch2018learning,klejch2019speaker}, and cross-accent speech recognition~\cite{winata2020learning}.

\paragraph{Code-Switching ASR} \citet{li2012code} introduces a statistical method to incorporate a linguistic theory into a code-switching speech recognition system, and \citet{adel2013recurrent,adel2013combination} explore syntactic and semantic features on recurrent neural networks (RNNs). ~\citet{baheti2017curriculum} adapts effective curriculum learning by training a network with monolingual corpora of two languages, and subsequently training on code-switched data. 
\citet{pratapa2018language} and \citet{lee2019cslm} propose to use methods to generate artificial code-switching data using a linguistic constraint. \citet{winata2018code} proposes to leverage syntactic information to improve the identification of the location of code-switching points, and improve the language model performance. Finally \citet{garg2018cslm} and \citet{winata2019code} propose new neural-based methods using SeqGAN and pointer-generator (Pointer-Gen) to generate diverse synthetic code-switching sentences that are sampled from the real code-switching data distribution.

\section{Meta-Transfer Learning}
We aim to effectively transfer knowledge from source domains to a specific target domain. We denote our model by $f_{\theta}$ with parameters $\theta$. Our model accepts a set of speech inputs $X = \{x_1,\ldots,x_n\}$ and generates a set of utterances $Y = \{y_1,\ldots,y_m\}$. The training involves a set of speech datasets in which each dataset is treated as a task $\mathcal{T}_i$. Each task is distinguished as either a source $\mathscr{D}_{src}$ or target task $\mathscr{D}_{tgt}$. For each training iteration, we randomly sample a set of data as training $\mathcal{D}^{tra}$, and a set of data as validation $\mathcal{D}^{val}$. In this section, we present and formalize the method.
\begin{algorithm}[t]
{\selectfont
\caption{Meta-Transfer Learning}
\label{alg1}
\textbf{Require:} $\mathscr{D}_{src}$, $\mathscr{D}_{tgt}$\\
\textbf{Require:} $\alpha, \beta$: step size hyperparameters
\begin{algorithmic}[1]
\State Randomly initialize $\theta$
\While{not done}
  \State Sample batch data $\mathcal{D}^{tra} \sim (\mathscr{D}_{src}, \mathscr{D}_{tgt})$, 
  
  $\mathcal{D}^{val} \sim \mathscr{D}_{tgt}$
  \For{\textbf{all} $\mathcal{D}^{tra}_{\mathcal{T}_i} \in \mathcal{D}^{tra}$}
      \State Evaluate $\nabla_{\theta}\mathcal{L}_{\mathcal{D}^{tra}_{\mathcal{T}_i}}(f_{\theta})$ using $\mathcal{D}^{tra}_{\mathcal{T}_i}$
      \State Compute adapted parameters with \phantom . \phantom .\phantom .\phantom .\phantom .\phantom .\phantom .\phantom .\phantom .\phantom .\phantom .\phantom .\phantom .gradient descent:
       \phantom . \phantom .\phantom .\phantom .\phantom .\phantom .\phantom .\phantom .\phantom .\phantom .\phantom .\phantom .\phantom .
       \phantom . \phantom .\phantom .\phantom .\phantom .\phantom .\phantom .\phantom .\phantom .\phantom .\phantom .\phantom .\phantom .
       \phantom .\phantom .\phantom .\phantom .\phantom .\phantom .\phantom .\phantom .\phantom .\phantom .\phantom .
       $\theta ^ { \prime }_{\mathcal{T}_i} = \theta - \alpha \nabla_{\theta}\mathcal{L}_{\mathcal{D}^{tra}_{\mathcal{T}_i}}(f_{\theta})$
\EndFor
  \State $\theta \leftarrow \theta-\beta\sum_{i} \nabla_{\theta}\mathcal{L}_{\mathcal{D}^{val}}\left(f_{\theta_{\mathcal{T}_i}^{\prime}}\right)$  
  
\EndWhile
\end{algorithmic}
}
\end{algorithm}
\subsection{Setup} To facilitate the model to achieve a good generalization on the code-switching data, we sample the source dataset $\mathscr{D}_{src}$ from monolingual English $(en)$ and Chinese $(zh)$ and code-switching $(cs)$ corpora, and choose the target dataset $\mathscr{D}_{tgt}$ only from the code-switching corpus. The code-switching data samples between $\mathscr{D}_{src}$ and $\mathscr{D}_{tgt}$ are disjoint. In this case, we exploit the meta-learning update using meta-transfer learning to acquire knowledge from the monolingual English and Chinese corpora, and optimize the learning process on the code-switching data. Then, we slowly fine-tune the trained model to become closer to the code-switching domain by avoiding aggressive updates that can push the model to a worse position.

\subsection{Meta-Transfer Learning Algorithm}

Our approach extends the meta-learning paradigm to adapt knowledge learned from source domains to a specific target domain. This approach captures useful information from multiple resources to the target domain, and updates the model accordingly. Figure~\ref{fig:model} presents the general idea of meta-transfer learning. The goal of the meta-transfer learning is not to focus on generalizing to all tasks, but to focus on acquiring crucial knowledge to transfer from monolingual resources to the code-switching domain. As shown in Algorithm 1, for each adaptation step on $\mathcal{T}_i$, we compute updated parameters $\theta ^ { \prime }_{\mathcal{T}_i}$ via stochastic gradient descent (SGD) as follows:
\begin{align}
    \theta ^ { \prime }_{\mathcal{T}_i} = \theta - \alpha \nabla_{\theta}\mathcal{L}_{\mathcal{D}^{tra}_{\mathcal{T}_i}}(f_{\theta}),
\end{align}
where $\alpha$ is a learning hyper-parameter of the inner optimization. Then, a cross-entropy loss $\mathcal{L}_{\mathcal{D}^{val}}$ is calculated from a learned model upon the generated text given the audio inputs on the target domain:
\begin{align}
    \mathcal{L}_{\mathcal{D}^{val}} = - \sum_{\mathcal{D}^{val}\sim \mathscr{D}_{tgt}} \log p(y_t|x_t;\theta'_{\mathcal{T}_i}).
\end{align}
We define the objective as follows:
\begin{align}
    &\min_{\theta}\sum_{\mathcal{D}_{\mathcal{T}_i}^{tra}, \mathcal{D}^{val}}\mathcal{L}_{\mathcal{D}^{val}}(f_{\theta_{\mathcal{T}_i}'}) = \\
    &\sum_{\mathcal{D}_{\mathcal{T}_i}^{tra}, \mathcal{D}^{val} }\mathcal{L}_{\mathcal{D}^{val}}(f_{\theta-\alpha \nabla_{\theta} \mathcal{L}_{D_{\mathcal{T}_i}^{tra}}(f_{\theta})}),
\end{align}
where $\mathcal{D}_{\mathcal{T}_i}^{tra}\sim (\mathscr{D}_{src},\mathscr{D}_{tgt})$ and $\mathcal{D}^{val}\sim \mathscr{D}_{tgt}$. We minimize the loss of the $f_{\theta'_{\mathcal{T}_i}}$ upon $\mathcal{D}^{val}$. Then, we apply gradient descent on the meta-model parameter $\theta$ with a $\beta$ meta-learning rate.

\section{Code-Switched Speech Recognition}
\subsection{Model Description}
We build our speech recognition model on a transformer-based encoder-decoder~\cite{dong2018speech,winata2019code}. The encoder employs VGG \cite{simonyan2014very}
to learn a language-agnostic audio representation and generate input embeddings. The decoder receives the encoder outputs and applies multi-head attention to the decoder input. We apply a mask into the decoder attention layer to avoid any information flow from future tokens. During the training process, we optimize the next character prediction by shifting the transcription by one. Then, we generate the prediction by maximizing the log probability of the sub-sequence using beam search.

\subsection{Language Model Rescoring}

To further improve the prediction, we incorporate \textit{Pointer-Gen LM}~\cite{winata2019code} in a beam search process to select the best sub-sequence scored using the softmax probability of the characters. We define $P(Y)$ as the probability of the predicted sentence. We add the pointer-gen language model  $p_{lm}(Y)$ to rescore the predictions. We also include word count \textit{wc(Y)} to avoid generating very short sentences. $P(Y)$ is calculated as follows:
\begin{equation}
P(Y) = \alpha P(Y|X) + \beta p_{lm}(Y)  + \gamma \sqrt{wc(Y)},
\end{equation}
where $\alpha$ is the parameter to control the decoding probability, $\beta$ is the parameter to control the language model probability, and $\gamma$ is the parameter to control the effect of the word count.

\begin{table}[!t]
\centering
\resizebox{0.42\textwidth}{!}{
\begin{tabular}{@{}rccc@{}}
\hline
\multicolumn{1}{|l|}{} & \multicolumn{1}{c|}{\textbf{Train}} & \multicolumn{1}{c|}{\textbf{Dev}} & \multicolumn{1}{c|}{\textbf{Test}} \\ \hline
\multicolumn{1}{|r|}{\# Speakers}               & \multicolumn{1}{c|}{138} & \multicolumn{1}{c|}{8} & \multicolumn{1}{c|}{8} \\
\multicolumn{1}{|r|}{\# Duration (hr)} & \multicolumn{1}{c|}{100.58} & \multicolumn{1}{c|}{5.56} & \multicolumn{1}{c|}{5.25} \\ 
\multicolumn{1}{|r|}{\# Utterances} & \multicolumn{1}{c|}{90,177} & \multicolumn{1}{c|}{5,722} & \multicolumn{1}{c|}{4,654} \\ 
\multicolumn{1}{|r|}{CMI} & \multicolumn{1}{c|}{0.18}& \multicolumn{1}{c|}{0.22} & \multicolumn{1}{c|}{0.19} \\ 
\multicolumn{1}{|r|}{SPF} & \multicolumn{1}{c|}{0.15} & \multicolumn{1}{c|}{0.19} & \multicolumn{1}{c|}{0.17} \\ \hline
\end{tabular}
}
\caption{Data statistics of SEAME Phase II. CMI and SPF represents \textit{code mixing index} and \textit{switch-point fraction}, respectively.}
\label{data-statistics-phase-2}
\end{table}

\begin{table}[!t]
\centering
\resizebox{0.49\textwidth}{!}{
\begin{tabular}{l|c}
\hline
 \textbf{Model} & \textbf{CER} \\ \hline
\citet{winata2019code} & 32.76\% \\
\hspace{3mm}+ Pointer-Gen LM & 31.07\% \\ \hline
Only \textit{CS} & 34.51\% \\ \hline
Joint Training (\textit{EN} + \textit{ZH}) & 98.29\% \\
\hspace{3mm}+ Fine-tuning & 31.22\% \\ \hline
Joint Training (\textit{EN} + \textit{CS}) & 34.77\% \\
Joint Training (\textit{ZH} + \textit{CS}) & 33.93\% \\
Joint Training (\textit{EN} + \textit{ZH} + \textit{CS}) & 32.87\% \\
\hspace{3mm}+ Fine-tuning & 31.90\% \\
\hspace{3mm}+ Pointer-Gen LM & 31.74\% \\ \hline
Meta-Transfer Learning (\textit{EN} + \textit{CS}) & 32.35\% \\
Meta-Transfer Learning (\textit{ZH} + \textit{CS}) & 31.57\% \\
Meta-Transfer Learning (\textit{EN} + \textit{ZH} + \textit{CS}) & 30.30\% \\
\hspace{3mm}+ Fine-tuning & 29.99\% \\
\hspace{3mm}+ Pointer-Gen LM & \textbf{29.30\%} \\ \hline
\end{tabular}
}
\caption{Results of the evaluation in CER, a lower CER is better. \textbf{\textit{Meta-Transfer Learning}} is more effective in transferring information from monolingual speech.}
\label{results}
\end{table}

\section{Experiments and Results}

\subsection{Dataset}
We use
SEAME Phase II, a conversational
English-Mandarin Chinese code-switching speech corpus that consists of spontaneously spoken interviews and conversations~\cite{SEAME2015}. 
The data statistics and code-switching metrics, such as code mixing index (CMI)~\cite{gamback2014measuring} and switch-point fraction~\cite{pratapa2018language} are depicted in  Table~\ref{data-statistics-phase-2}. For monolingual speech datasets, we use HKUST~\cite{liu2006hkust} as the monolingual Chinese dataset, and Common Voice~\cite{ardila2019common} as the monolingual English dataset.\footnote{We downloaded the CommonVoice version 1 dataset from https://voice.mozilla.org/.} We use 16 kHz audio inputs and up-sample the HKUST data from 8 to 16 kHz.

\subsection{Experiment Settings}
Our transformer model consists of two encoder layers and four decoder layers with a hidden size of 512, an embedding size of 512, a key dimension of 64, and a value dimension of 64. The input of all the experiments uses spectrogram, computed with a 20 ms window and shifted every 10 ms. Our label set has 3765 characters and includes all of the English and Chinese characters from the corpora, spaces, and apostrophes. We optimize our model using Adam and start the training with a learning rate of 1e-4. We fine-tune our model using SGD with a learning rate of 1e-5, and apply an early stop on the validation set. We choose $\alpha=1$, $\beta=0.1$, and $\gamma=0.1$. We draw the sample of the batch randomly with a uniform distribution every iteration. 

We conduct experiments with the following approaches: \textbf{(a)} only \textit{CS}, \textbf{(b)} joint training on \textit{EN} + \textit{ZH}, \textbf{(c)} joint training on \textit{EN} + \textit{ZH} + \textit{CS}, and \textbf{(d)} meta-transfer learning. Then, we apply fine-tuning \textbf{(b)}, \textbf{(c)}, and \textbf{(d)} models on \textit{CS}. We apply LM rescoring on our best model. We evaluate our model using beam search with a beam width of 5 and maximum sequence length of 300. The quality of our model is measured using character error rate (CER).

\subsection{Results}

The results are shown in Table~\ref{results}. Generally, adding monolingual data \textit{EN} and \textit{ZH} as the training data is effective to reduce error rates. There is a significant margin between \textbf{only \textit{CS}} and \textbf{joint training} (1.64\%) or \textbf{meta-transfer learning} (4.21\%). According to the experiment results, \textbf{meta-transfer learning} consistently outperforms the joint-training approaches. This shows the effectiveness of meta-transfer learning in language adaptation. 

The fine-tuning approach helps to improve the performance of trained models, especially on the joint training (\textit{EN} + \textit{ZH}). We observe that joint training (\textit{EN} + \textit{ZH}) without fine-tuning cannot predict mixed-language speech, while joint training on \textit{EN} + \textit{ZH} + \textit{CS} is able to recognize it. However, according to Table~\ref{comparison}, adding a fine-tuning step badly affects the previous learned knowledge (e.g., \textit{EN}: 11.84\% $\rightarrow$ 63.85\%, \textit{ZH}: 31.30\% $\rightarrow$ 78.07\%). Interestingly, the model trained with meta-transfer learning does not suffer catastrophic forgetting even without focusing the loss objective to learn both monolingual languages. As expected, joint training on \textit{EN} + \textit{ZH} + \textit{CS} achieves decent performance on all tasks, but it does not optimally improve \textit{CS}.

\begin{figure}[!t]
    \centering
    \includegraphics[scale=0.37]{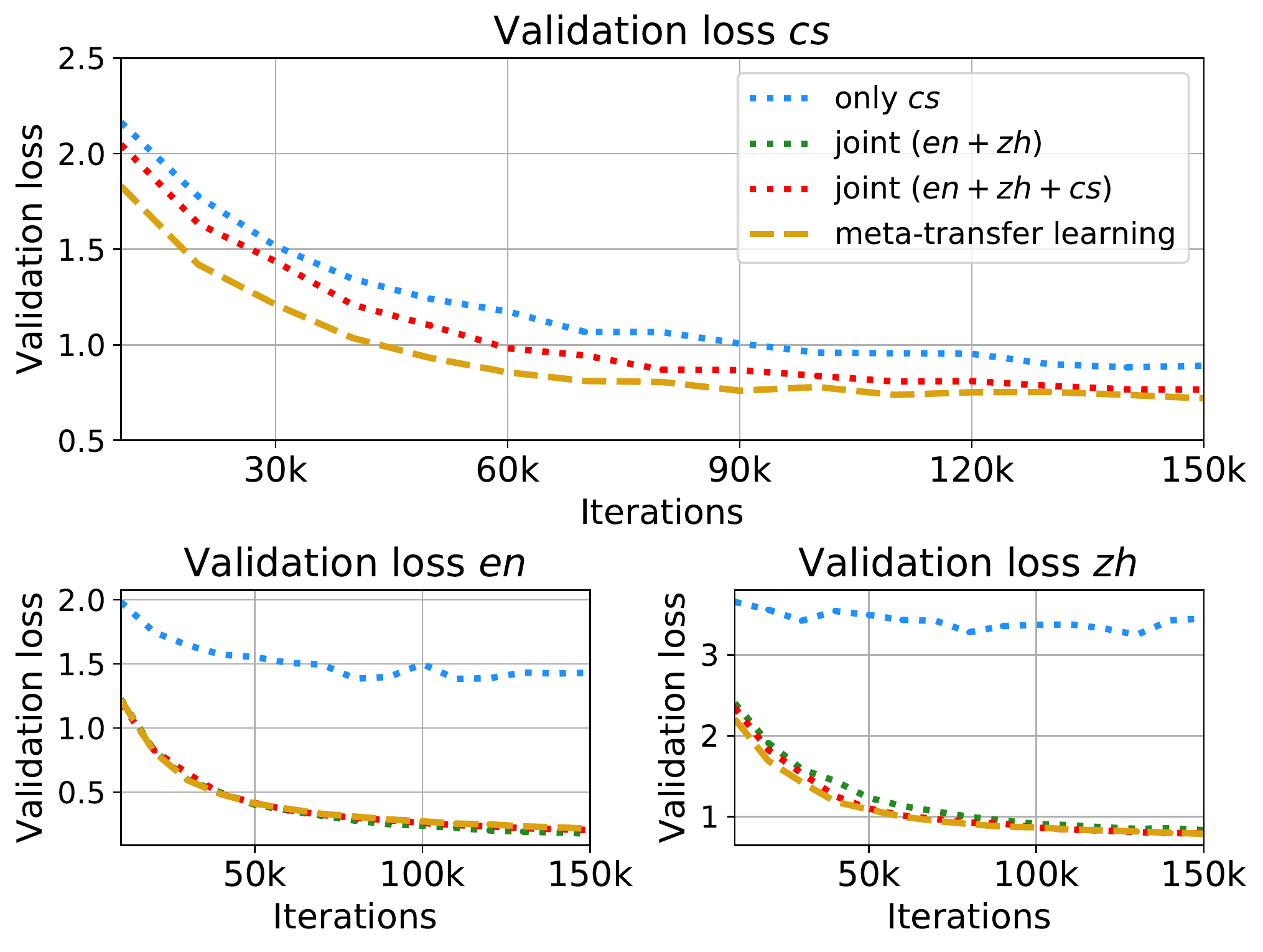}
    \caption{Validation loss per iteration. \textbf{Top:} validation loss on \textit{CS} data, 
    (joint (\textit{EN} + \textit{ZH}) is omitted because it is higher than the range), 
    \textbf{bottom left:} validation loss on \textit{EN} data, \textbf{bottom right:} validation loss on \textit{ZH} data.}
    \label{fig:validation_loss}
\end{figure}

\begin{table*}[!t]
\centering
\resizebox{0.71\textwidth}{!}{
\begin{tabular}{l|c|c|c}
\hline \textbf{Model} & \textbf{$\uparrow\Delta$ CS} & \textbf{$\downarrow$ EN} & \textbf{$\downarrow$ ZH} \\ \hline
Only \textit{CS} & - & 66.71\% & 99.66\% \\ \hline
Joint Training (\textit{EN} + \textit{ZH}) & -63.78\% & 11.84\% & 31.30\% \\
\hspace{3mm}+ Fine-tuning & 3.29\% & 63.85\% & 78.07\% \\  
Joint Training (\textit{EN} + \textit{ZH} + \textit{CS}) & 1.64\% & 13.88\% & 30.46\% \\
\hspace{3mm}+ Fine-tuning & 2.61\% & 57.56\% & 76.20\% \\  
\hline
Meta-Transfer Learning (\textit{EN} + \textit{ZH} + \textit{CS}) & \textbf{4.21\%} & \textbf{16.22\%} & \textbf{31.39\%} \\ \hline
\end{tabular}
}
\caption{Performance on monolingual English CommonVoice test set (\textit{EN}) and HKUST test set (\textit{ZH}) in CER. $\Delta$ CS denotes the improvement on SEAME test set (\textit{CS}) relative to the baseline model (Only \textit{CS}).}
\label{comparison}
\end{table*}

The language model rescoring using Pointer-Gen LM improves the performance of the meta-transfer learning model by choosing more precise code-switching sentences during beam search. Pointer-Gen LM improves the performance of the model, and outperforms the model trained only in \textit{CS} by 5.21\% and previous state-of-the-art by 1.77\%.

\paragraph{Convergence Rate} Figure~\ref{fig:validation_loss} depicts the dynamics of the validation loss per iteration on \textit{CS}, \textit{EN}, and \textit{ZH}. As we can see from the figure, meta-transfer learning is able to converge faster than only \textit{CS} and joint training, and results in the lowest validation loss. For the validation losses on \textit{EN} and \textit{ZH}, both joint training (\textit{EN} + \textit{ZH} + \textit{CS}) and meta-transfer learning achieve a similar loss in the same iteration, while only \textit{CS} achieves a much higher validation loss. This shows that meta-transfer learning is not only optimized on the code-switching domain, but it also preserves the generalization ability to monolingual domains, as depicted in Table~\ref{comparison}.


\subsection{Language Modeling Task}

We further evaluate our meta-transfer learning approach on a language model task. We simply take the transcription of the same datasets and build a 2-layer LSTM-based language model following the model configuration in~\citet{winata2019code}. To further improve the performance, we apply fine-tuning with an SGD optimizer by using a learning rate of 1.0, and decay the learning rate by 0.25x for every epoch without any improvement on the validation performance. To prevent the model from over-fitting, we add an early stop of 5 epochs.

\begin{table}[!t]
\centering
\resizebox{0.49\textwidth}{!}{
\begin{tabular}{l|c|c}
\hline
\multirow{1}{*}{\textbf{Model}} & \multicolumn{1}{c|}{\textbf{valid}} & \multirow{1}{*}{\textbf{test}} \\ \hline
Only \textit{CS}$^{\ddagger}$ & 72.89 & 65.71  \\ \hline
Joint Training (\textit{EN} + \textit{ZH} + \textit{CS}) & \multicolumn{1}{c|}{70.99} & 63.73 \\ 
\hspace{3mm}+ Fine-tuning & 69.66 & 62.73 \\ \hline
Meta-Transfer Learning (\textit{EN} + \textit{ZH} + \textit{CS}) & \multicolumn{1}{c|}{68.83} & 62.14 \\ 
\hspace{3mm}+ Fine-tuning & \textbf{68.71} & \textbf{61.97} \\ \hline
\end{tabular}
}
\caption{Results on the language modeling task in perplexity. $^{\ddagger}$ the results are from~\citet{winata2019code}.}
\label{lm-results}
\end{table}

As shown in Table~\ref{lm-results}, the meta-transfer learning approach outperforms the joint-training approach. We find a similar trend for the language model task results to the speech recognition task where meta-transfer learning without additional fine-tuning performs better than joint training with fine-tuning. Compared to our baseline model (Only \textit{CS}), meta-transfer learning is able to reduce the test set perplexity by 3.57 points (65.71 $\rightarrow$ 62.14), and the post fine-tuning step reduces the test set perplexity even further, from 62.14 to 61.97.


\section{Conclusion}
We propose a novel method, \textit{meta-transfer learning}, to transfer learn on a code-switched speech recognition system in a low-resource setting by judiciously extracting information from high-resource monolingual datasets. Our model recognizes individual languages and transfers them so as to better recognize mixed-language speech by conditioning the optimization objective to the code-switching domain. Based on experimental results, our training strategy outperforms joint training even without adding a fine-tuning step, and it requires less iterations to converge. 

In this paper, we have shown that our approach can be effectively applied to both speech processing and language modeling tasks. Finally, we will explore further the generability of our meta-transfer learning approach to more downstream multilingual tasks in our future work.

\section*{Acknowledgments}
This work has been partially funded by ITF/319/16FP and MRP/055/18 of the Innovation Technology Commission, the Hong Kong SAR Government, and School of Engineering Ph.D. Fellowship Award, the Hong Kong University of Science and Technology, and RDC 1718050-0 of EMOS.AI.

\bibliography{acl2020}
\bibliographystyle{acl_natbib}

\end{document}